\documentclass[10pt,twocolumn,letterpaper]{article}

\usepackage{cvpr}              

\usepackage[dvipsnames]{xcolor}

\definecolor{cvprblue}{rgb}{0.21,0.49,0.74}
\usepackage[pagebackref,breaklinks,colorlinks,citecolor=cvprblue]{hyperref}
\usepackage{multirow}
\usepackage{float}

\def\paperID{349} 
\def\confName{3DV\xspace}
\def\confYear{2025\xspace}

\title{Open-Vocabulary Semantic Part Segmentation of 3D Human}

\author{Keito Suzuki\textsuperscript{*,1} \quad Bang Du\textsuperscript{*,1} \quad Girish Krishnan\textsuperscript{*} \quad Kunyao Chen\textsuperscript{$\dagger$} \quad Runfa Blark Li\textsuperscript{*} \quad Truong Nguyen\textsuperscript{*}\\
\textsuperscript{*}University of California, San Diego \quad \textsuperscript{$\dagger$}Qualcomm\\
{\tt\small \{k3suzuki, b7du, gikrishnan, kuc017, runfa, tqn001\}@ucsd.edu}
}

\begin{document}
\twocolumn[{%
\renewcommand\twocolumn[1][]{#1}%
\maketitle
\begin{figure}[H]
\hsize=\textwidth
\centering
\includegraphics[width=0.95\textwidth]{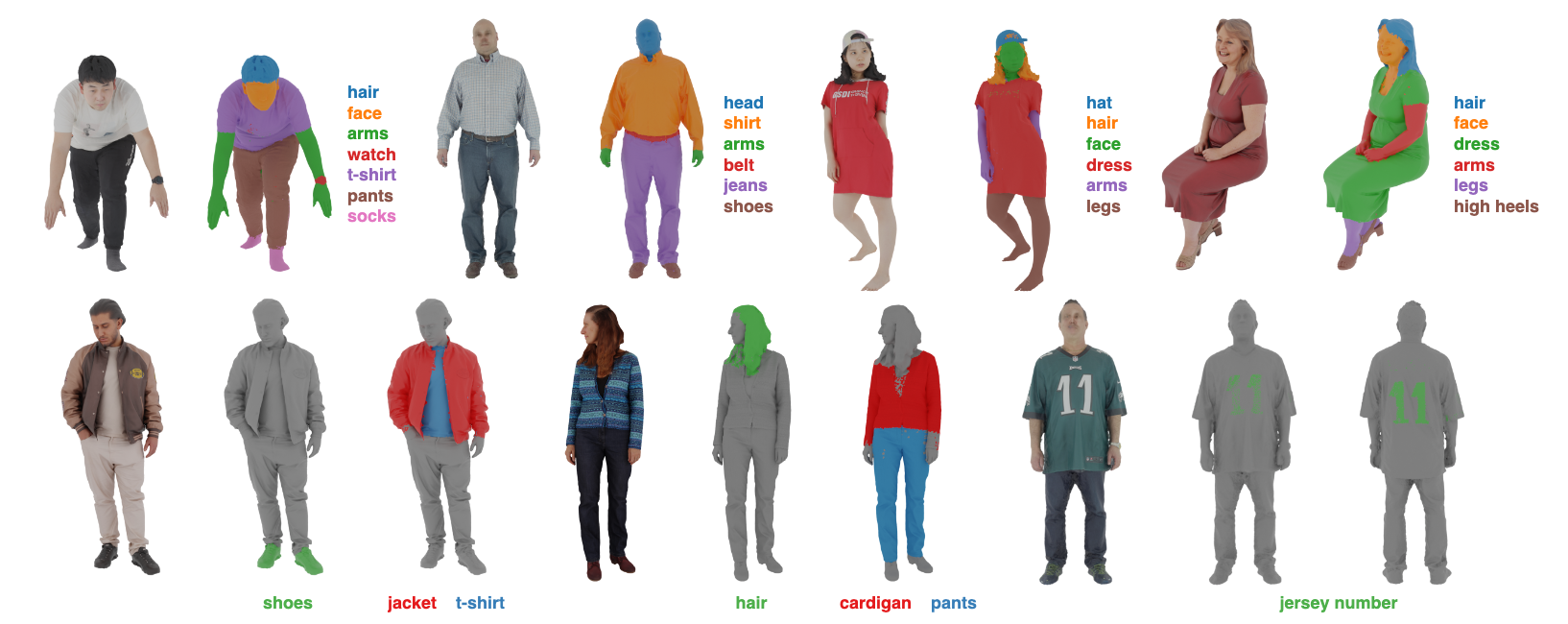}
\caption{We propose the first open-vocabulary method for the segmentation of 3D human. It infers 3D segmentation by rendering multi-view images and leveraging pre-trained vision-language models. The figure displays the input text prompts and the corresponding segmentation results for 3D humans from various datasets. Our method supports arbitrary queries and generates non-overlapping masks in the 3D model. See Figure \ref{Fig.promptable_seg2} and Figure \ref{Fig.visual} for more results. 
}
\label{Fig.intro}
\end{figure}
}]

\begin{abstract}
3D part segmentation is still an open problem in the field of 3D vision and AR/VR. Due to limited 3D labeled data, traditional supervised segmentation methods fall short in generalizing to unseen shapes and categories. Recently, the advancement in vision-language models' zero-shot abilities has brought a surge in open-world 3D segmentation methods. While these methods show promising results for 3D scenes or objects, they do not generalize well to 3D humans. In this paper, we present the first open-vocabulary segmentation method capable of handling 3D human. Our framework can segment the human category into desired fine-grained parts based on the textual prompt. We design a simple segmentation pipeline, leveraging SAM to generate multi-view proposals in 2D and proposing a novel HumanCLIP model to create unified embeddings for visual and textual inputs. Compared with existing pre-trained CLIP models, the HumanCLIP model yields more accurate embeddings for human-centric contents. We also design a simple-yet-effective MaskFusion module, which classifies and fuses multi-view features into 3D semantic masks without complex voting and grouping mechanisms. The design of decoupling mask proposals and text input also significantly boosts the efficiency of per-prompt inference. Experimental results on various 3D human datasets show that our method outperforms current state-of-the-art open-vocabulary 3D segmentation methods by a large margin. In addition, we show that our method can be directly applied to various 3D representations including meshes, point clouds, and 3D Gaussian Splatting.
\end{abstract}

\footnotetext[1]{Equal Contribution.}    
\section{Introduction}
\label{sec:intro}

The advancements in 3D technologies have led to an increased demand for automated analysis of 3D shapes. Among the related tasks, 3D part segmentation plays a pivotal role in supporting a wide spectrum of applications, including robotics and AR/VR. 

With the introduction of deep neural networks \cite{qi2017pointnet, qi2017pointnet++, thomas2019kpconv, xu2021paconv, zhao2021point} and labeled 3D datasets \cite{wu20153d, chang2015shapenet}, 3D part segmentation has seen remarkable progress in recent years through supervised training. Nonetheless, creating 3D datasets is expensive and time-consuming. Compared with image data, current 3D part-annotated datasets are orders of magnitude smaller in scale. Within the limited 3D data, the human category represents only a tiny fraction. Existing human parsing methods have been trained to segment clothed data \cite{bhatnagar2019multi, musoni2023gim3d, antic2024close} or underlying body parts \cite{bogo2014faust,takmaz20233d}, but they fall short of generalizing to unseen models and classes. Thus, enabling machines with the ability to parse objects into semantic parts and generalize to new categories still remains difficult, especially for human-related data, which usually contain more complex geometry with richer semantic attributes than general 3D objects.

Recent developments in vision-language learning gave rise to many 2D image-based models \cite{radford2021learning, li2022grounded} with exceptional zero-shot generalization capabilities. Many works seek to transfer 2D knowledge to 3D through pre-trained image-language models. \cite{zhang2022pointclip, zhu2023pointclip, abdelreheem2023satr, liu2023partslip, zhou2023partslip++} leverage these models through multi-view rendering and aggregate the information in 3D for the final segmentation result. Another line \cite{umam2024partdistill} focuses on distilling the information for a better 3D model. While these methods have shown promising improvements in object data, they have not exhibited the same quality of results on 3D human data.


In this paper, we aim to bring the open-vocabulary 3D part segmentation performance to human data. We introduce the first framework for 3D human parsing that semantically segments whatever parts you want and supports various 3D representations, including meshes, point clouds, and 3D Gaussians \cite{kerbl20233d}. Inspired by \cite{cheng2021per}, we formulate the segmentation task as a mask classification problem. Firstly, we generate class-agnostic instance mask proposals on rendered images through a pre-trained 2D segmentation model, SAM \cite{kirillov2023segment}. Secondly, we propose a novel HumanCLIP model that encodes each mask into embeddings within the CLIP feature space. Compared to the vanilla CLIP model \cite{radford2021learning}, HumanCLIP produces more accurate text-aligned embeddings for human-related cases, enhancing the precision of the final segmentation. Since mask proposals are class-agnostic and independent between views, we propose a novel MaskFusion module that simultaneously classifies the semantic labels given the text prompt and fuses the multi-view inconsistent masks to generate 3D semantic segmentation for the input. It decouples the mask proposal step from reliance on text prompts, thereby enhancing inference efficiency. In summary, our contributions mainly include:
\begin{itemize}
    \item We introduce the first open-vocabulary framework that segments fine-grained parts for 3D humans.
    \item We present a HumanCLIP model capable of extracting discriminative CLIP embeddings for human-centric data. 
    \item We propose a MaskFusion module that simultaneously classifies semantic labels and converts multi-view 2D proposals into 3D segmentation, which significantly enhances inference efficiency. 
    \item Our framework shows state-of-the-art performance on five 3D human datasets and shows compatibility with various 3D representations including 3D Gaussian Splatting.
\end{itemize}

\section{Related Works}
\label{sec:related_works}

\begin{figure*}[t]
    \centering
    \includegraphics[width=.95\hsize]{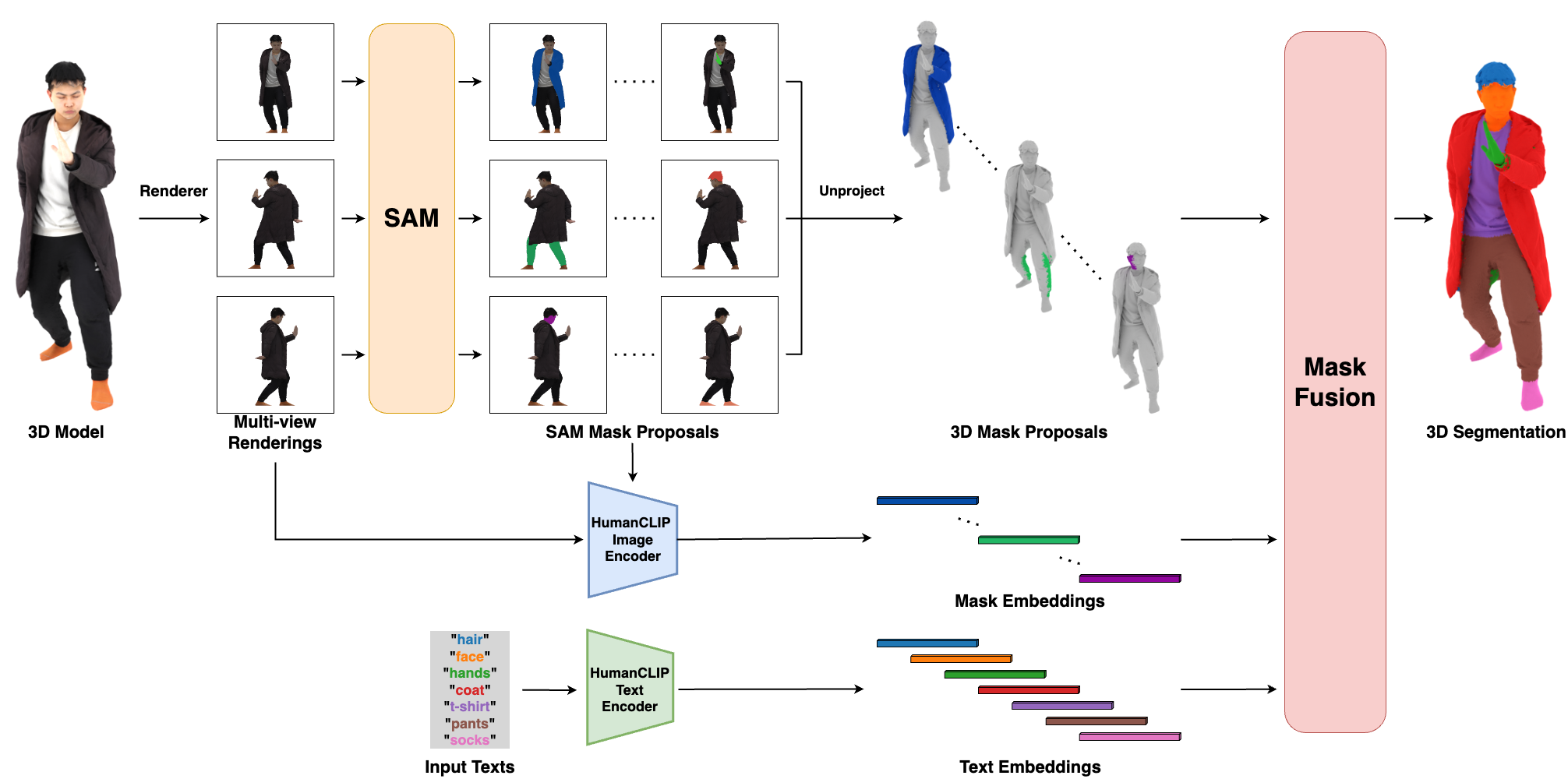}
    \caption{Overview of the proposed framework. Given a 3D human model, it is first rendered to obtain multi-view 2D images. The images are then fed to SAM to generate class-agnostic 2D masks and unprojected to obtain binary 3D masks. Additionally, each pair of image and 2D masks are fed to the human-centric mask-based text-aligned image encoder to obtain CLIP embeddings for each mask. Simultaneously, the input class texts are fed to the text encoder to obtain corresponding text embeddings. The 3D mask proposals, mask embeddings, and text embeddings are fed to the mask fusion module to obtain the final segmentation result.} 
    \label{Fig.model}
\end{figure*}

\subsection{3D Human Part Segmentation}

The 3D human part segmentation field is largely driven by the advancement of 3D neural networks as well as the labeled dataset. \cite{jertec2019using, ueshima2021training, musoni2023gim3d} train point cloud \cite{qi2017pointnet, qi2017pointnet++, thomas2019kpconv} or mesh segmentation networks \cite{sharp2022diffusionnet, hanocka2019meshcnn, dong2023laplacian2mesh} through direct supervision or leveraging the unclothed human parametric templates or physical simulation of garments. \cite{takmaz20233d} curates synthetic data to boost the performance on body parts. Datasets such as SIZER \cite{tiwari2020sizer}, MGN \cite{bhatnagar2019multi}, and CTD \cite{chen2021tightcap}, provide coarse clothing labels from 3D scans of clothed humans, but having only three categories and less variations of poses limit their applications. \cite{antic2024close} presents Close-D which consists of 18 garment categories. To the best of our knowledge, it is the most comprehensive dataset up-to-date. However, due to the still limited size, none of these methods show generalizable ability and cannot segment labels outside of the predefined taxonomy. 


\subsection{Open-Vocabulary 3D Segmentation}
In recent years, large vision language models \cite{radford2021learning, jia2021scaling} have grown popular due to their ability to perform zero-shot recognition. As a result, many works have incorporated these models \cite{ding2022decoupling, li2022languagedriven, ghiasi2022scaling, liang2023open} to conduct open-vocabulary 2D segmentation. To transfer the knowledge into 3D, some methods \cite{zhang2022pointclip, huang2022clip2point, zhu2023pointclip} apply CLIP to depth maps for zero-shot object classification and segmentation. For scene-scale data, CLIP2Scene \cite{chen2023clip2scene} and CLIP$^2$ \cite{zeng2023clip2} train an additional 3D encoder with a contrastive loss. Although these models have shown their effectiveness on general 3D scenes or objects, we find that they do not work well for 3D humans. As one of the most important categories, a framework tailored for open-vocabulary 3D human parsing is demanded. 

\noindent\textbf{CLIP Embeddings.}
CLIP \cite{radford2021learning} is one of the most widely used vision-language models in both 2D and 3D open-world segmentation approaches. Due to being trained with natural images, the vanilla CLIP does not perform well on special input subsets, such as masked images or fashion images. Many approaches try to adapt CLIP models to new tasks. \cite{chia2022contrastive, cartella2023openfashionclip} presents fine-tuned CLIP on fashion data. \cite{liang2023open} propose the Mask-adapted CLIP. AlphaCLIP \cite{sun2024alpha} adds an additional alpha channel as input so that it composes both regional and global information for better understanding. While AlphaCLIP provides better embeddings than CLIP for a region of interest, we observe that it is still inadequate to distinguish human body parts and garments.


\section{Proposed Method}

\subsection{Overview}

The overview of the proposed framework is shown in Figure \ref{Fig.model}. We assume a point-based 3D shape with size $P$ as the input. Given the 3D human model and $K$ semantic text prompts, the goal is to parse the 3D human into segments that semantically correspond to the input prompts.

Inspired by recent achievements in 2D and 3D segmentation methods \cite{cheng2022masked, ding2022decoupling, zhou2022lmseg, schult2023mask3d, takmaz20233d}, we formulate the semantic segmentation task as mask classification, originated from MaskFormer \cite{cheng2021per}. To bridge 3D data with 2D pre-trained models, we render the input from $V$ predefined camera views. Segment-Anything-Model (SAM) \cite{kirillov2023segment} is leveraged to generate mask proposals for each view (Section \ref{mask_proposals}). We introduce the novel HumanCLIP model, which encodes these proposals into embeddings within the unified CLIP feature space (Section \ref{mask_embeddings}). To lift 2D labels into 3D, it is usually required to assign ``super points'' and have carefully designed voting and grouping. In this work, we present a simple MaskFusion module. It takes HumanCLIP encoded text prompts and simultaneously performs classification and multi-view aggregation without the need for complex operations (Section \ref{mask_fusion}). Note that the generation of mask proposals and embeddings is performed just once per model. Subsequently, segmentation can be executed in just a few milliseconds per prompt, significantly enhancing efficiency compared to previous methods.



\subsection{Multi-view Mask Proposals}
\label{mask_proposals}
We choose SAM \cite{kirillov2023segment} to generate mask proposals on multi-view rendered images. SAM demonstrates remarkable zero-shot capabilities in image segmentation. From the predefined camera poses, we render the 3D human into $V$ RGB images $I_i \in \mathbb{R}^{H\times W \times 3}$ where $i \in [1, V]$ is the index of the view. Each image $I_i$ is then independently fed into SAM in the "segment everything" mode to generate $N_i$ class-agnostic overlapping masks: 
\begin{equation}
    [m_{i, 1}^{2D}, ..., m_{i, N_i}^{2D}] = SAM(I_i)
\end{equation}
where $m_{i, j}^{2D}$ is the $j$-th 2D mask generated by SAM from the $i$-th view.
This results in a total of $N=\sum_{i=1}^V N_i$ binary 2D masks at a varying granularity of whole, part, and subpart. 

Each 2D mask $m_{i, j}^{2D}$ is then unprojected to 3D using the camera parameters of view $i$ to construct the corresponding 3D proposal $m_{i, j}^{3D}$.

\subsection{HumanCLIP Encoding}
\label{mask_embeddings}
We propose HumanCLIP to generate proposal embeddings with size $C$, where $C$ is the embedding dimension of a CLIP model. The unified image-text feature space of CLIP allows the framework to perform open-vocabulary mask classification. Each 2D mask $m_{i, j}^{2D}$ with its corresponding image $I_i$ is fed to the image encoder to get the proposal embedding $q_{i, j} \in \mathbb{R}^{C}$. 

\begin{figure}[t]
    \centering
    \includegraphics[width=.9\hsize]{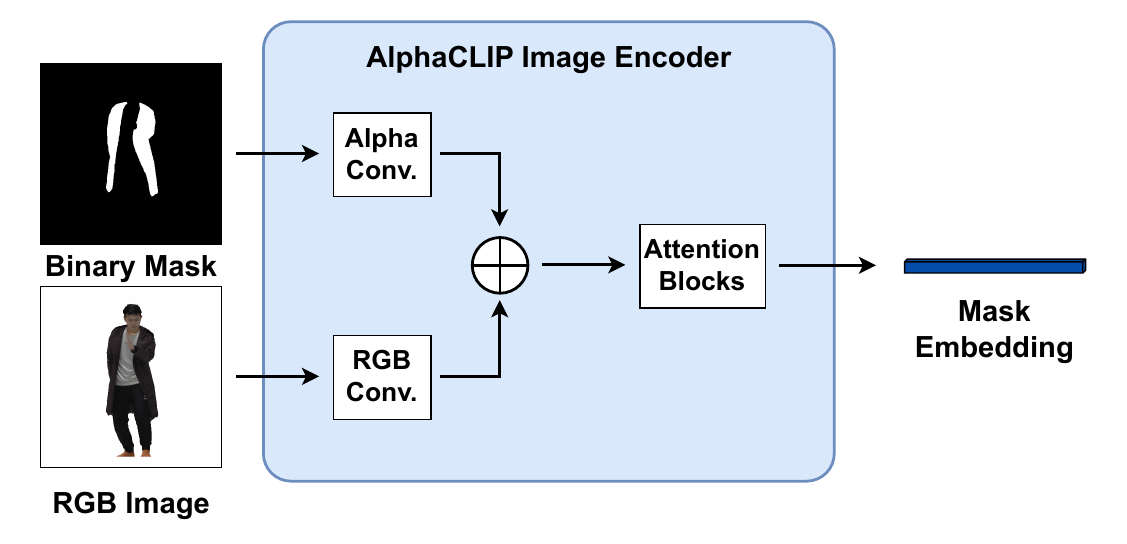}
    \caption{AlphaCLIP Image Encoder.} 
    \label{Fig.alphaclip}
\end{figure}

\begin{figure}[t]
    \centering
    \includegraphics[width=.98\hsize]{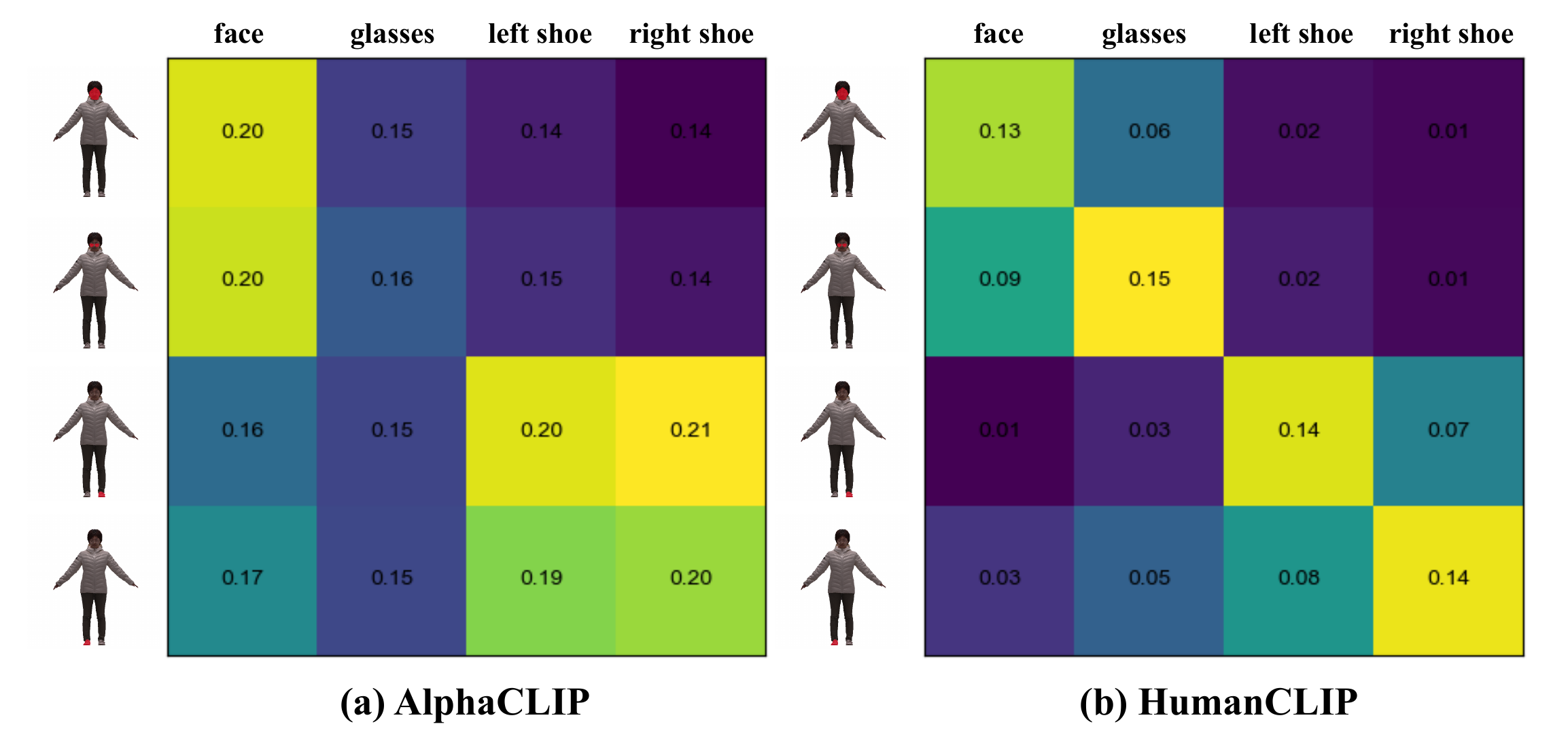}
    \caption{Comparison between (a) pre-trained AlphaCLIP and (b) the proposed HumanCLIP. The plots show the cosine similarity between the embedding of the masked region corresponding to \textit{face}, \textit{glasses}, \textit{left shoe}, and \textit{right shoe} and their text embeddings.} 
    \label{Fig.finetune}
\end{figure}

It is well-discussed that the vanilla pre-trained CLIP encoder does not perform well on specialty-formed inputs \cite{liu2023partslip,liang2023open}, including masked and cropped images. Moreover, masking or cropping an image results in the loss of crucial contextual information, which is essential to the understanding of the specific area in an image. Therefore, we adopt the design of AlphaCLIP \cite{sun2024alpha} to build our image encoder. As shown in Figure \ref{Fig.alphaclip}, the encoder accepts an additional alpha channel as input, which highlights the region of interest on the original rendered images. The input mask is processed with a parallel convolution layer to the RGB image and combined to go through a series of attention blocks to produce the final mask embedding in CLIP feature space. To further mitigate the domain gap, we finetune the encoder on a dataset of over 1.3 million RGBA region-text pairs with human-centric contents. We visualize the image-text alignment before and after fine-tuning in Figure \ref{Fig.finetune}. The pre-trained AlphaCLIP model fails to provide well-aligned embeddings for small parts such as the glasses as well as to distinguish left and right parts. The proposed HumanCLIP model generates more discriminative mask embeddings, facilitating the downstream classification tasks. 

\subsection{Region-Text Pair Generation}
\label{sec:regiontext}
\begin{figure}[t]
    \centering
    \includegraphics[width=.98\hsize]{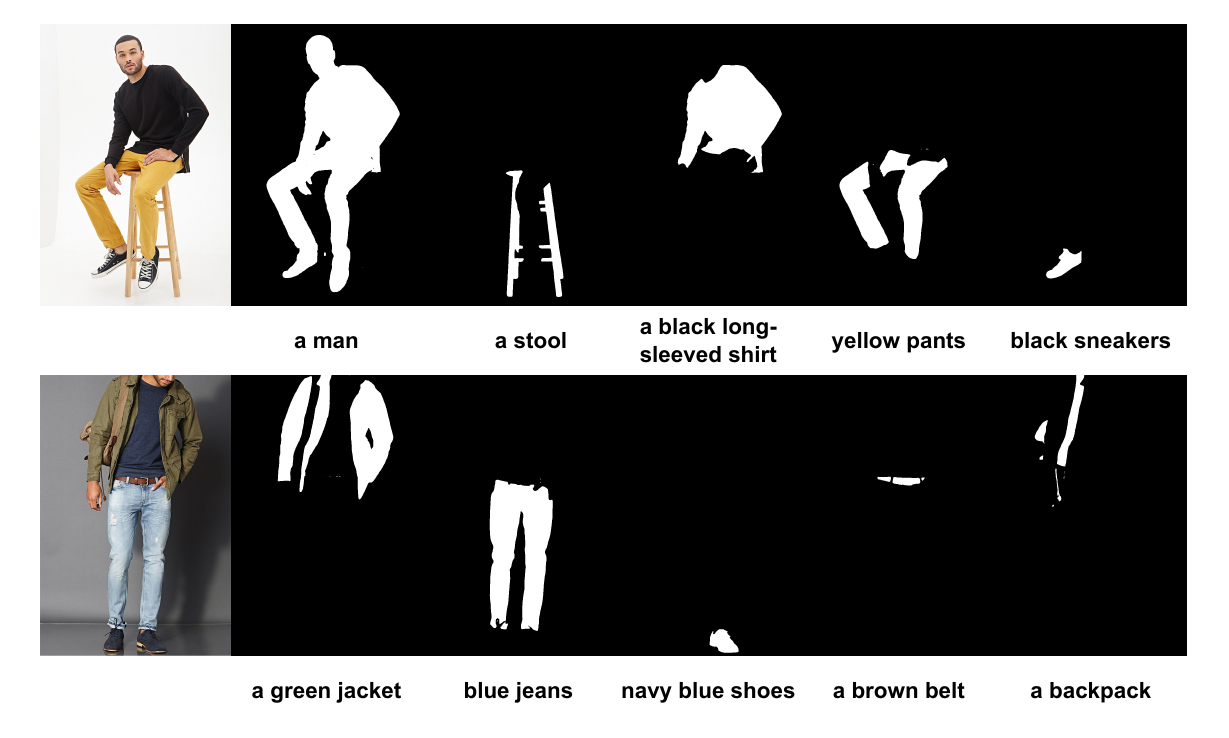}
    \caption{Example of mask-caption pairs generated by utilizing KOSMOS-2 and SAM.} 
    \label{Fig.example_masks}
\end{figure}

To tailor the image encoder for processing human-centric data, we finetune the model with region-text pair data. A straightforward method to acquire this data is utilizing 2D human segmentation datasets, where segmentation maps and category names directly form region-text pairs. Although efficient, this method yields less diverse masks and less informative captions. Therefore, we devise a pipeline to augment the training data. We source images from LIP \cite{gong2017look}, ATR \cite{liang2015deep}, DeepFashion \cite{liu2016deepfashion}, and CIHP \cite{gong2018instance} datasets and employ KOSMOS-2 \cite{peng2023kosmos} and SAM \cite{kirillov2023segment} to automatically generate masks and corresponding captions for these images. An example of the generated pairs is depicted in Figure \ref{Fig.example_masks}. Compared with the original labels, it provides more descriptive captions and introduces novel masks for objects that humans typically interact with, such as as `a stool'. Further details of the data generation process are presented in the supplementary. 

\subsection{3D Semantic Segmentation}
\label{mask_fusion}
To obtain the segmentation result with the desired semantic labels, our pipeline accepts $K$ text prompts corresponding to the labels per inference. These texts are fed to the HumanCLIP text encoder to obtain CLIP text embeddings $\mathbf{W} \in \mathbb{R}^{K \times C}$. Then, the proposed MaskFusion module semantically classifies and synthesizes multi-view embeddings into 3D segmentation masks. Specifically, we utilize the correlations between the text embeddings and the mask embeddings to build correspondences and fuse the independent 3D masks. 

Recap the $N$ 3D mask proposals generated in Section. \ref{mask_proposals}. The proposals and their embeddings are stacked to get $\mathbf{M} \in \mathbb{R}^{P \times N}$ and $\mathbf{Q} \in \mathbb{R}^{N \times C}$ respectively. We compute the classification logits $\mathbf{P} \in \mathbb{R}^{N \times K}$ by taking the cosine similarity between each mask embedding and each text embedding:
\begin{equation}
    \mathbf{P}_{n, k} = \frac{\mathbf{Q}_n\cdot \mathbf{W}_k}{||\mathbf{Q}_n||||\mathbf{W}_k||}
\end{equation}
It is used to guide the grouping of raw masks, which are class-agnostic and inconsistent across views. 

In the final step, for each 3D point, we aggregate the class scores from the associated masks to get the final 3D segmentation result $\mathbf{Y} \in \mathbb{R}^{P \times K}$.
$Y$ is computed as the simple weighted average of 3D masks $\mathbf{M}$ based on the classification logits $\mathbf{P}$:
\begin{equation}
    \mathbf{Y} = \mathbf{M} \times \mathbf{P}
\end{equation}

We decouple the procedures for mask proposal and for text classification. Therefore, it is not guaranteed that each text input is valid for the 3D model. To ensure that only existed classes are segmented in the final result, we set a threshold $\tau$ on the final segmentation logits. If the maximum logits of a point fall below $\tau$, the point is attributed to an `other' class. 
\section{Experiments}
\subsection{Implementation Details}

\begin{table}
    \centering
    \caption{Statistics of region-text pairs used for HumanCLIP training. Each mask is accompanied by a descriptive caption. }
    \label{tab:humanclip_data}
    \resizebox{.98\columnwidth}{!}{
    \begin{tabular}{c|ccc|c}
       Dataset & Images & Original Masks & Generated Masks & Total Masks \\ \hline
       LIP & 30462 & 173578 & 91505 & 265083 \\
       ATR & 17706 & 175604 & 87698 & 263302 \\
       DeepFashion & 12701 & 100632 & 43404 & 144036 \\
       CIHP & 28280 & 647072 & 63616 & 710688 \\ \hline
       \textbf{HumanCLIP} & \textbf{89149} & \textbf{1096886} & \textbf{286223} & \textbf{1383109} 
    \end{tabular}
    }
\end{table}

\noindent\textbf{Segment Anything Model.} 
When applying SAM in our framework for both 3D segmentation and training data generation, we adopt the ViT-H model checkpoint. To create the mask proposals, we feed a rendered image of resolution $512 \times 512$ in the ``segment everything" mode where we sample 64 points along each side of the image.

\noindent\textbf{HumanCLIP.}
We initialize the HumanCLIP image encoder with the AlphaCLIP ViT-L/14 checkpoint, which is pre-trained on GRIT-20m dataset \cite{peng2023kosmos} with an image resolution of $224\times224$. The model is then finetuned on the curated HumanCLIP dataset, in which the images combines four 2D human parsing datasets: LIP \cite{gong2017look}, ATR \cite{liang2015deep}, DeepFashion \cite{liu2016deepfashion}, and CIHP \cite{gong2018instance}. We utilize both the ground truth segmentation maps and the augmented region-text pairs described in Section \ref{sec:regiontext}, resulting in approximately 1.38 million RGBA-caption pairs for training. The distribution of images and masks across these datasets is detailed in Table \ref{tab:humanclip_data}. We keep the text encoder frozen and finetune the image encoder for a total of 3 epochs with a batch size of 18.

\subsection{Effectiveness of HumanCLIP}

\begin{figure}[]
    \centering
    \includegraphics[width=.98\hsize]{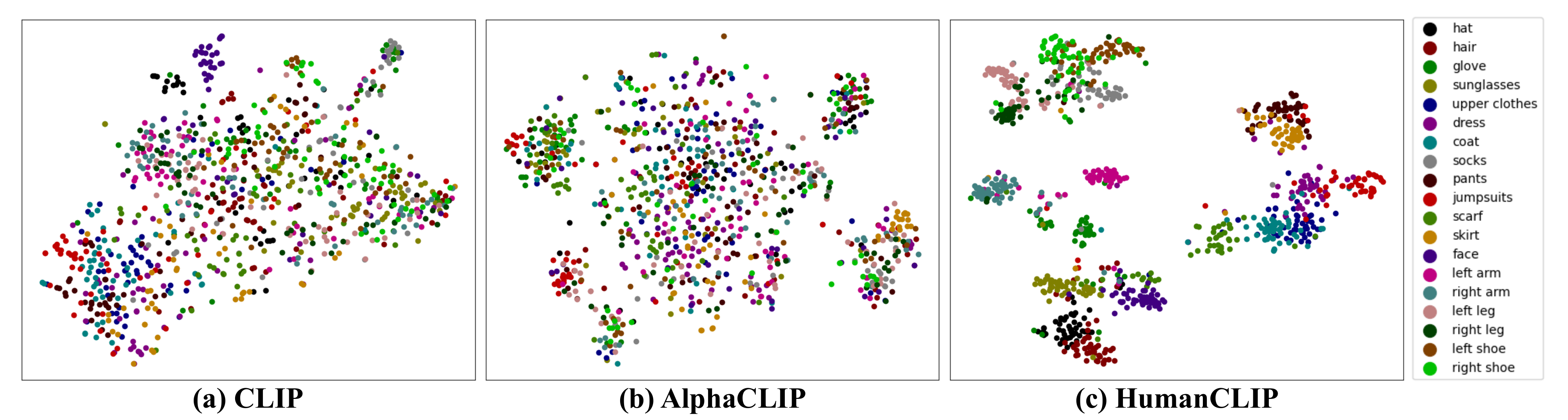}
    \caption{Comparison of (a) CLIP, (b) AlphaCLIP and (c) HumanCLIP's t-SNE \cite{van2008visualizing} visualizations of the mask embeddings for categories in the LIP dataset.} 
    \label{Fig.tsne}
\end{figure}

\noindent
\textbf{Embedding Space.}
In Figure \ref{Fig.tsne} we draw the t-SNE \cite{van2008visualizing} projection of the mask embeddings extracted by CLIP, AlphaCLIP, and HumanCLIP on the LIP dataset. 
For CLIP and AlphaCLIP, significant overlap among embeddings of different categories is observed, complicating accurate class distinction based on text. In contrast, our proposed HumanCLIP forms more well-defined clusters for each class, enhancing the discriminativeness of the features.

\noindent
\textbf{Mask Classification.}
To further assess the effectiveness of the proposed HumanCLIP model in accurately embedding human parts, we conducted a mask classification task comparing it with the vanilla CLIP and pre-trained AlphaCLIP models. In this task, we first feed each ground truth image and mask to the image encoder to extract the mask embedding and then compute the cosine similarity with the text embedding for each class. For the CLIP models, the image segment cropped from the ground truth mask is input to the encoder to generate image embeddings. Classification is determined by the highest similarity score. The results, displayed in Table \ref{tab:mask_classification}, show classification accuracy on the LIP \cite{gong2017look} and CCP \cite{yang2014clothing} datasets, which feature 19 and 54 classes respectively. The results indicate that both CLIP and AlphaCLIP models underperform, attributed to their lack of training on specific human parts data. AlphaCLIP shows slight improvements over CLIP as it provides better mask-wise embeddings.
It is evident that the proposed HumanCLIP significantly outperforms both models in correctly classifying each mask based on the extracted embeddings.

\begin{table}[]
    \centering
    \caption{Comparison of CLIP, AlphaCLIP, and HumanCLIP on mask classification accuracy of the LIP \cite{gong2017look} and CCP \cite{yang2014clothing} dataset.}
    \label{tab:mask_classification}
    \resizebox{0.45\columnwidth}{!}{
        \begin{tabular}{c|cc}
        Model & LIP & CCP \\ \hline
        CLIP & 22.12 & 21.75\\
        AlphaCLIP & 27.86 & 22.51 \\
        HumanCLIP & \textbf{79.98} & \textbf{52.96}
    \end{tabular}
    }
\end{table}

\begin{table*}
    \centering
    \caption{Comparison with open-set 3D segmentation methods. OA, mAcc, and mIoU are the overall accuracy, mean class accuracy, and mean Intersection over Union respectively. For each metric, a higher value is better. The best results are shown in \textbf{bold}.}
    \label{tab:quantitative}
    \resizebox{\textwidth}{!}{\begin{tabular}{c|ccc|ccc|ccc|ccc|ccc|ccc}
        \multirow{2}{*}{Model} & \multicolumn{3}{c|}{MGN} & \multicolumn{3}{c|}{SIZER} & \multicolumn{3}{c|}{CTD} & \multicolumn{3}{c|}{THuman2.0} & \multicolumn{3}{c|}{PosedPro} & \multicolumn{3}{c}{Average} \\ \cline{2-19}
        & OA & mAcc & mIoU & OA & mAcc & mIoU & OA & mAcc & mIoU & OA & mAcc & mIoU & OA & mAcc & mIoU & OA & mAcc & mIoU\\ \cline{1-19}
        PointCLIP V2 & 21.41 & 24.15 & 13.31 & 44.80 & 34.42 & 20.92 & 11.06 & 13.68 & 5.94 & 3.46 & 14.51 & 1.77 & 6.67 & 13.53 & 2.33 & 17.48 & 20.06 & 8.85\\
        SATR & 84.72 & 77.30 & 67.17 & 82.00 & 81.97 & 67.38 & 78.55 & 86.20 & 64.98 & 56.05 & 34.31 & 20.60 & 51.61 & 43.28 & 22.43 & 70.59 & 64.61 & 48.51 \\
        PartSLIP & 90.03 & 86.53 & 78.63 & 84.94 & 82.18 & 70.79 & 75.11 & 71.20 & 55.46 & 77.46 & 44.94 & 33.70 & 70.38 & 34.61 & 24.80 & 74.58 & 63.89 & 52.68 \\
        PartSLIP++ & 91.41 & 87.98 & 81.14 & 86.63 & 83.49 & 72.93 & 80.73 & 76.06 & 62.36 & 82.00 & 49.82 & 38.96 & 70.80 & 35.07 & 24.99 & 82.31 & 66.48 & 56.08\\
        Ours & \textbf{94.61} & \textbf{95.03} & \textbf{88.78} & \textbf{91.24} & \textbf{90.73} & \textbf{82.55} & \textbf{93.29} & \textbf{93.21} & \textbf{83.36} & \textbf{89.88} & \textbf{69.40} & \textbf{54.50} & \textbf{80.23} & \textbf{46.37} & \textbf{37.27} & \textbf{89.85} & \textbf{78.95} & \textbf{69.29}\\
     \end{tabular}}
\end{table*}

\begin{figure}[t]
    \centering
    \includegraphics[width=.95\hsize]{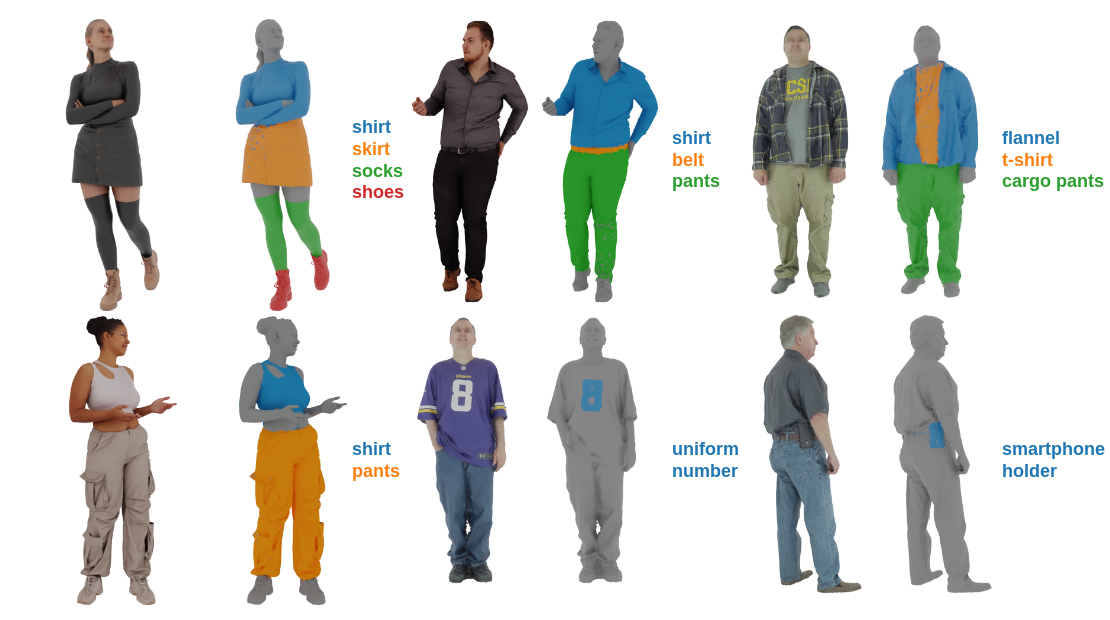}
    \caption{Examples of promptable segmentation.} 
    \label{Fig.promptable_seg2}
\end{figure}

\subsection{Comparison with Open-Vocabulary 3D Segmentation Methods}
\noindent
\textbf{Methods.}
To the best of our knowledge, there is no method dedicated to open-vocabulary 3D human parsing. Hence, we conduct comparisons with four general 3D segmentation approaches: PointCLIP v2 \cite{zhu2023pointclip}, SATR \cite{abdelreheem2023satr}, PartSLIP \cite{liu2023partslip}, and PartSLIP++ \cite{zhou2023partslip++}. PointCLIP v2 applies CLIP to multi-view depth maps for zero-shot 3D classification, part segmentation, and object detection. SATR applies the GLIP model \cite{li2022grounded} to rendered images and aggregates the multi-view bounding predictions for each prompt to yield a segmented mesh. It shows capabilities under unclothed human setting. PartSLIP also applies GLIP but for low-shot point cloud segmentation. This is then enhanced by PartSLIP++ which incorporates SAM and an EM algorithm.

\begin{figure*}[t]
    \centering
    \includegraphics[width=.85\hsize]{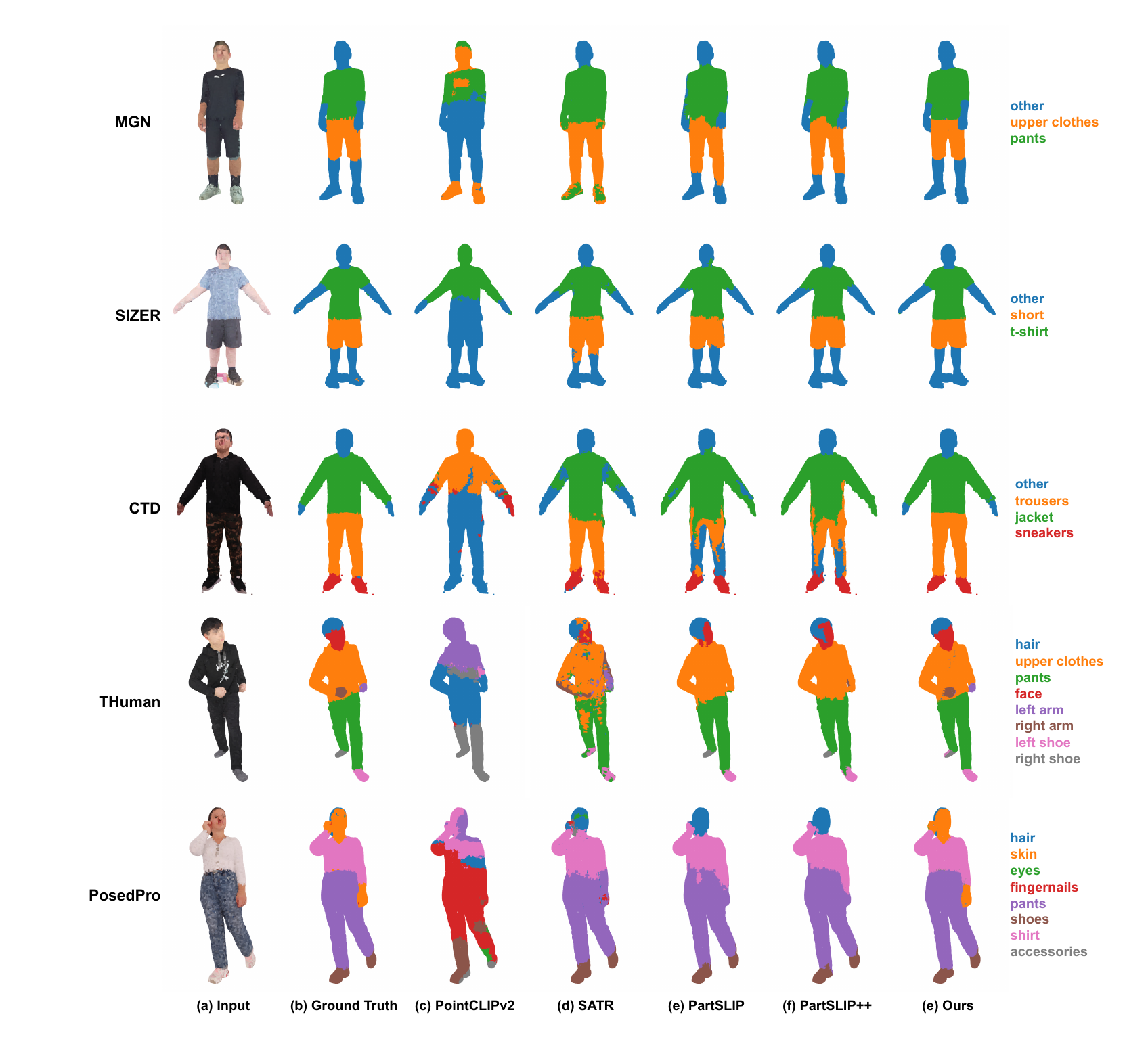}
    \caption{Qualitative analysis of segmentation results with (c) PointCLIPv2, (d) SATR, (e) PartSLIP, (f), PartSLIP++ on the 3D scans from MGN, SIZER, CTD, THuman, and PosedPro datasets.} 
    \label{Fig.visual}
\end{figure*}

\noindent
\textbf{Datasets.}
For quantitative evaluation, we benchmark the models on five labeled 3D human datasets: MGN \cite{bhatnagar2019multi}, SIZER \cite{tiwari2020sizer}, CTD \cite{chen2021tightcap}, THuman2.0 \cite{tao2021function4d}, and Posed Pro \cite{RenderPeople}. These dataset contain a total of 3, 9, 12, 12, and 19 classes respectively.


\noindent
\textbf{Quantitative Comparison.}
In Table \ref{tab:quantitative}, we show the quantitative comparison of our model with the four open-set 3D segmentation methods. We first notice that PointCLIP v2 performs poorly across all of the datasets. Since they apply CLIP to rendered depth maps, which differs greatly from real-world images that it was originally trained on, PointCLIP v2 is unable to effectively transfer the zero-shot capabilities to 3D. Among the other three methods, performance is generally best on the MGN dataset, which has the fewest classes, and declines as the number of classes increases. Across all datasets, it is evident that our proposed framework outperforms by a large margin in all metrics.

\noindent
\textbf{Visual Comparison.}
In Figure \ref{Fig.visual}, we show the visual comparison of various methods on the same five 3D human datasets. Akin to the numerical results, PointCLIPv2 fails to generate reasonable segmentation results. SATR and PartSLIP are able to get a coarse segmentation: the boundaries between various segments are unclear. PartSLIP++ shows improved boundaries but still struggles with specific areas like ‘hair’ and ‘face’. In contrast, our method delivers the most precise segmentation results.


\subsection{Promptable Segmentation}

The design of our MaskFusion module allows users to segment whatever they want, as highlighted in Figure \ref{Fig.promptable_seg2}. Beyond conducting a full segmentation of the entire body, our framework can precisely segment only the user-specified items. It also effectively recognizes and accurately segments unseen categories, such as ``uniform number'' and ``smartphone holder''.


\subsection{Run-time Efficiency}
Rendering 3D data into multi-view images for processing has raised efficiency concerns in applications. Our approach, which decouples mask proposal generation from textual prompt processing, offers a significant efficiency advantage over previous methods. The mask embedding serves as an attribute of the 3D model, which can be generated beforehand. During the segmentation phase, only the text encoder and MaskFusion module are active. In contrast, for \cite{abdelreheem2023satr, liu2023partslip, zhou2023partslip++}, the GLIP model relies on text prompts to generate bounding boxes and masks, requiring the entire pipeline to be executed for each segmentation attempt.


Table \ref{tab:efficiency} displays the inference times for various methods when executed on a system equipped with a single 24GB RTX 4090 graphics card. We compare the inference time for a one-time only run, which executes all modules from start to finish, and the average time for 100 inferences of the same model, during which we reuse any pre-generated information where possible. For instance, we do not rerender multi-view images for the subsequent inferences. The results of our approach show a significant decrease in average cost as the number of inferences increases. The most time-consuming step in our framework is the ``segment everything" mode of SAM. However, this is only necessary once for multiple text inputs, making our model efficient for subsequent inferences. In scenarios common to the open-vocabulary setting, where there is a fixed amount of 3D assets but varying information is required based on user queries, our method offers considerable advantages. 



\begin{table}[t]
    \centering
    \caption{Comparison of inference times. We compare the average time cost in seconds assuming one-time inference only and 100 inference calls. }
    \label{tab:efficiency}
    \resizebox{\columnwidth}{!}{\begin{tabular}{c|ccccc}
         & PointCLIPv2 & SATR & PartSLIP & PartSLIP++ & Ours \\ \hline
        One-time Inference & 7.62 & 54.18 & 32.46 & 88.41 & 105.72\\
        Average Inference & 1.27  & 27.69 & 26.08 & 74.55 & 1.06\\
    \end{tabular}
    }
\end{table}

\begin{figure}[]
    \centering
    \includegraphics[width=.8\hsize]{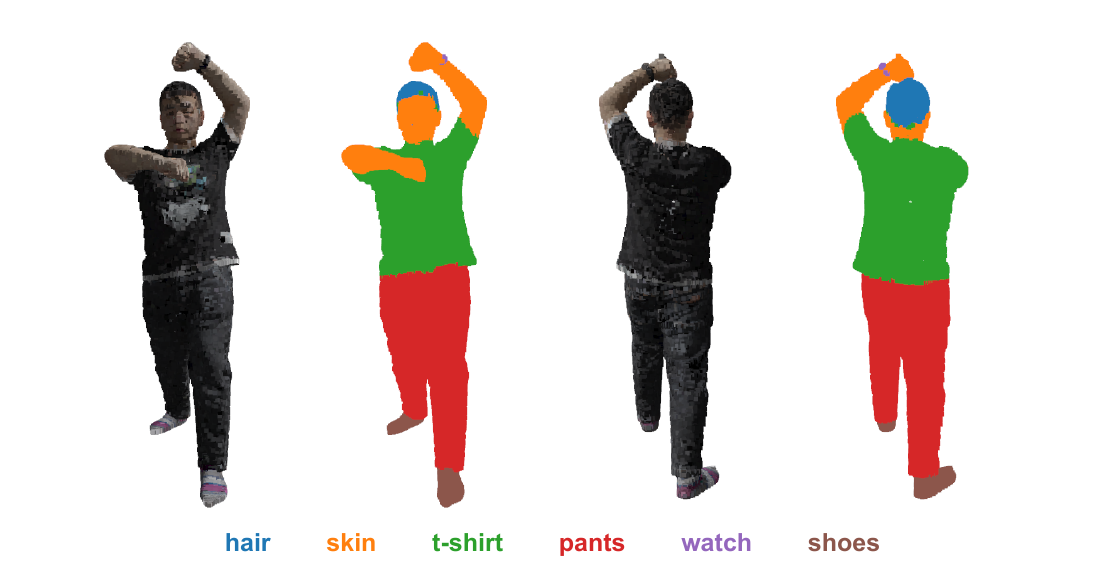}
    \caption{Segmentation of 3D Gaussian Splatting.} 
    \label{Fig.results_gaussian}
\end{figure}

\subsection{3D Gaussian Splatting Segmentation}

Our framework design is compatible with various point-based 3D representations, including the popular 3D Gaussian Splatting \cite{kerbl20233d} format. We follow the standard protocol to generate 3DGS for each model in the THuman2.0 dataset. As demonstrated in Figure \ref{Fig.results_gaussian} our method can generate reasonable segmentation results, making it a generalizable solution for segmenting 3DGS. This approach eliminates the need to optimize per-Gaussian semantic labels \cite{lan20232d} or high-dimensional features \cite{zhou2024feature} during the resource-intensive training stage.


\subsection{In-the-wild Segmentation}
\begin{figure}[]
    \centering
    \includegraphics[width=.9\hsize]{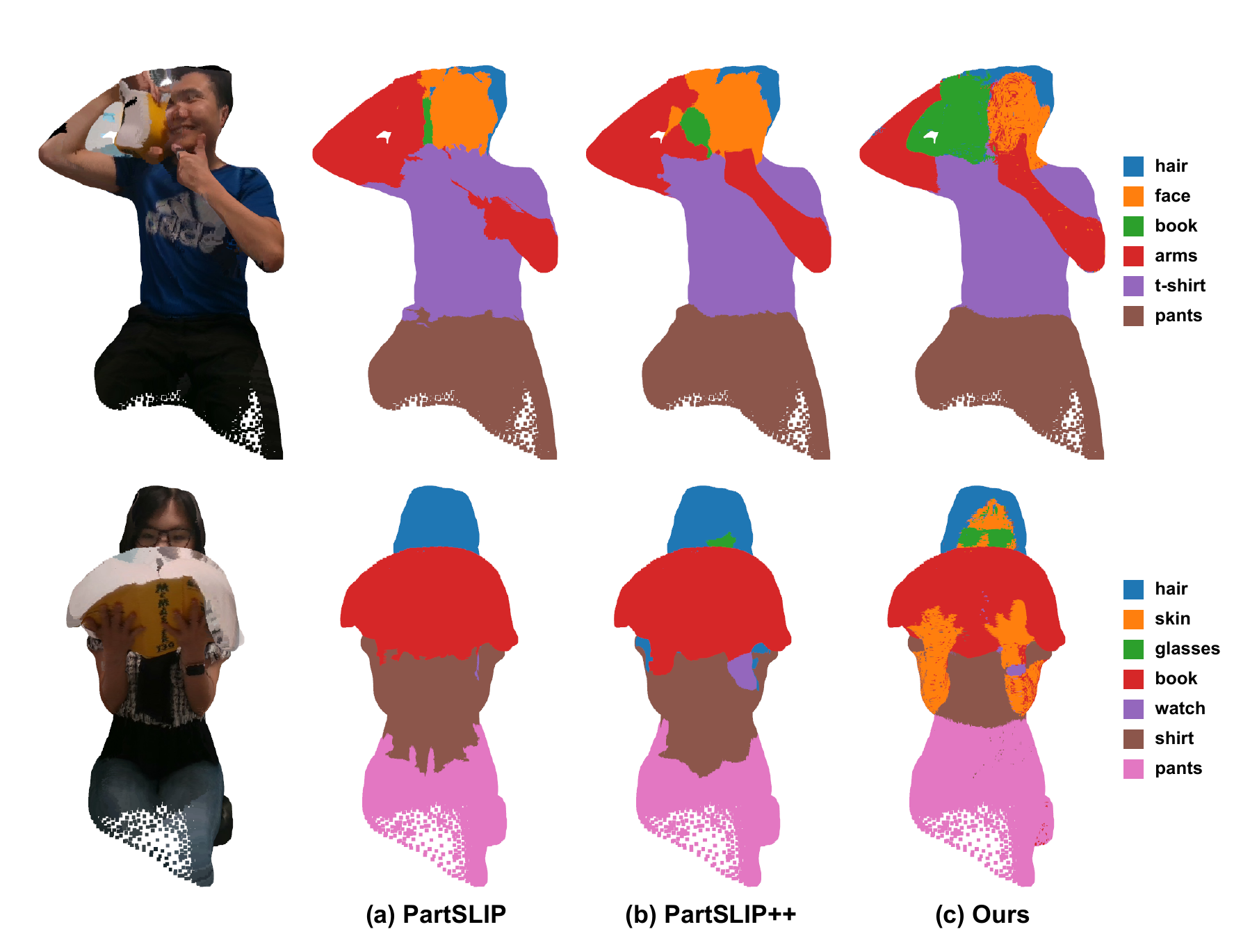}
    \caption{Visual comparison of (a) PartSLIP, (b) PartSLIP++, and (c) Ours on our in-the-wild point cloud dataset.} 
    \label{Fig.results_wild}
\end{figure}

Our method demonstrates no significant domain gap in real-world noisy scenarios. Figure \ref{Fig.results_wild} shows a visual comparison with PartSLIP and PartSLIP++ on two point clouds captured by consumer-level RGB-D sensors, where the surface is at relatively low definition and the model is incomplete. In the first example, all methods are able to accurately segment the clothing, but our method shows the best segmentation ability of the book. For the second example, both PartSLIP and PartSLIP++ are unable to segment the `skin' and `glasses' categories. In contrast, our method can accurately segment these classes as well as the small area for the watch. More details are explained in the supplementary. 

\section{Ablation Study}

\begin{figure}[]
    \centering
    \includegraphics[width=.95\hsize]{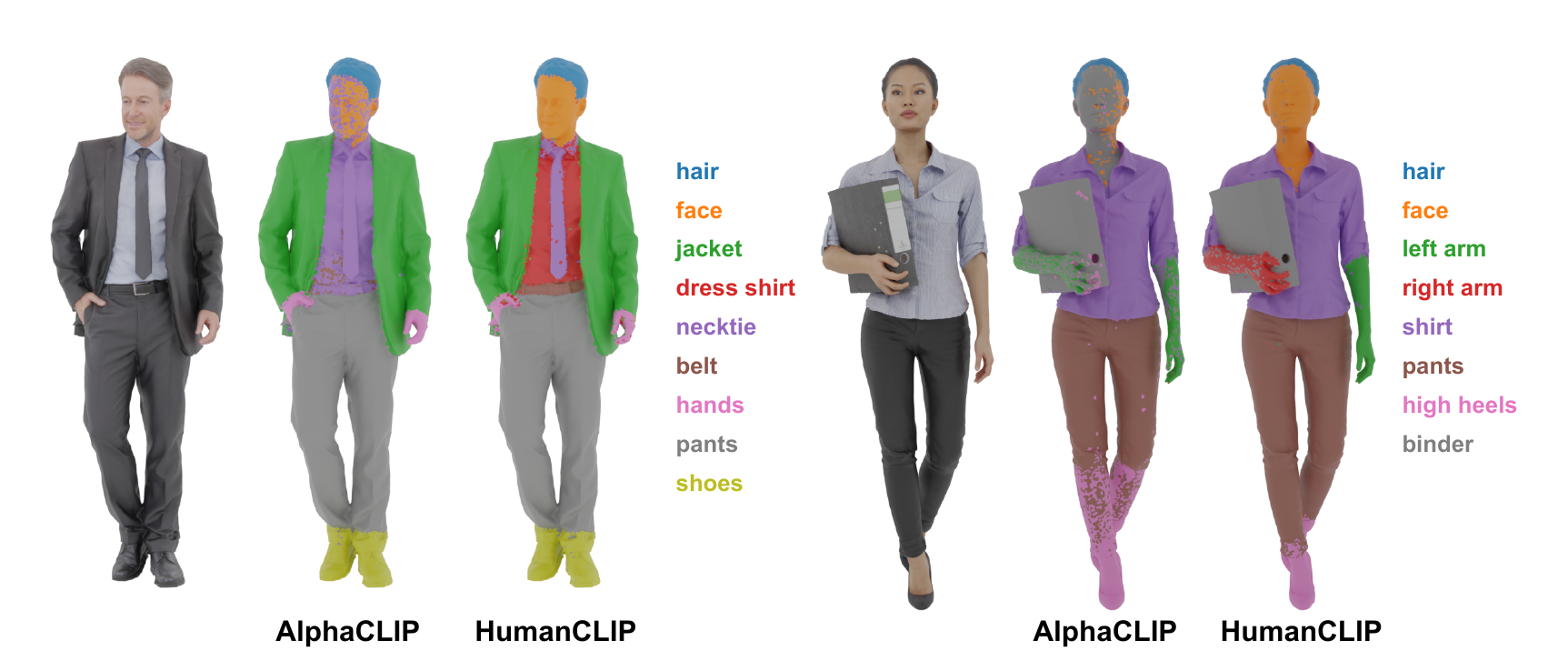}
    \caption{Visual comparison with AlphaCLIP in 3D human segmentation on the RenderPeople dataset.} 
    \label{Fig.compare_alphaclip}
\end{figure}

\noindent\textbf{HumanCLIP in Segmentation. }
Table \ref{tab:mask_classification} and Figure \ref{Fig.tsne} present the advantage of HumanCLIP in extracting discriminative embeddings on human-related mask-caption data. We further ablate the module within our framework to assess its contribution to overall performance. Figure \ref{Fig.compare_alphaclip} illustrates the segmentation results when replacing the HumanCLIP with pre-trained AlphaCLIP model. In the example on the left, while AlphaCLIP accurately segments areas such as the jacket, pants, and hands, it struggles to distinguish the inner layer of clothing. In the example on the right, AlphaCLIP results in noisy segmentation across the legs and the right side of the body. Conversely, using HumanCLIP enables precise segmentation of body parts and clothing, as well as neighboring objects like a binder.

\noindent\textbf{Number of views. }
To evaluate how the number of views affects the segmentation quality, we increase the number of views from 2 to 16 and compare the performance on the CTD dataset. The results are shown in Table \ref{tab:num_views}. We observe that as we increase the number of views, the segmentation quality improves as it can better mitigate noisy mask proposals from affecting the final result. However, more views also require more time to preprocess, so we select 8 views to strike a good balance between quality and efficiency. It’s important to note that all comparison methods utilize more than 8 views. The difference in the number of views does not confer an advantage to our method.

\begin{table}
    \centering
    \caption{Effect of the number of views on the segmentation quality.}
    \label{tab:num_views}
    \resizebox{0.7\columnwidth}{!}{\begin{tabular}{c|ccccc}
        \multirow{2}{*}{Metrics} & \multicolumn{5}{c}{Number of Views}\\ \cline{2-6}
        & 2 & 4 & 8 & 10 & 16 \\ \hline
         Accuracy & 91.48 & 92.31 & 93.29 & 93.44 & 93.71\\
         mAcc. & 91.43 & 92.09 & 93.21 & 93.49 & 93.61 \\
         mIoU. & 80.49 & 81.43 & 83.36 & 83.87 & 84.24\\
    \end{tabular}
    }
\end{table}

\noindent\textbf{Limitations.} One limitation of this method is the slow run-time for a single inference caused by applying SAM to every view. This can make it difficult to apply our method to dynamic 3D humans. Another limitation is that we have to manually adjust the threshold to conduct promptable segmentation for different text inputs. 
In future works, we plan on applying our method to generate labeled data to train a model that can efficiently compute in 3D space.

\section{Conclusion}
In this paper, we present the first open-vocabulary method for 3D human segmentation. We introduce a novel HumanCLIP model and a MaskFusion module, which efficiently transfer the knowledge from 2D pre-trained vision-language models to the segmentation of 3D human data. Our method can seamlessly conduct semantic segmentation based on arbitrary user-defined text queries. The experimental results show that our method outperforms existing open-vocabulary 3D segmentation methods on five 3D human datasets. Additionally, we show that our method can be directly applied to various 3D representations including points clouds, meshes, and 3D Gaussian Splatting. 

\noindent\textbf{Acknowledgments.} This research was supported by the Ministry of Trade, Industry and Energy (MOTIE) and Korea Institute for Advancement of Technology (KIAT) through the International Cooperative R\&D program in part (P0019797).
\newpage
{
    \small
    \bibliographystyle{ieeenat_fullname}
    \bibliography{main}
}
\clearpage
\setcounter{page}{1}
\maketitlesupplementary 
\appendix

\begin{figure}[]
    \centering
    \includegraphics[width=.95\hsize]{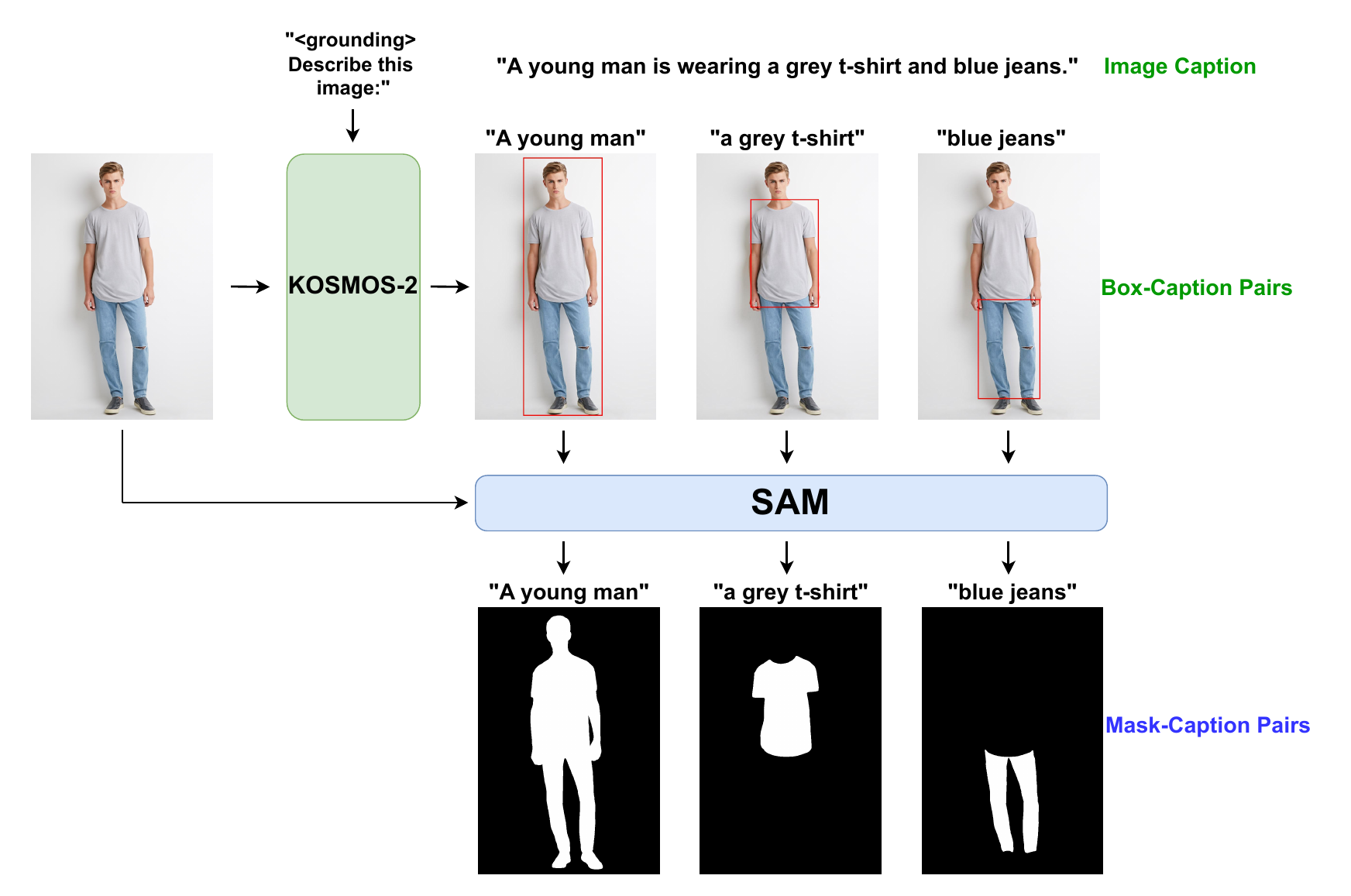}
    \caption{Data generation pipeline to create additional training data for HumanCLIP.} 
    \label{Fig.data_generation}
\end{figure}

\section{Data Generation Pipeline for HumanCLIP}
The pipeline used to automatically generate training data for HumanCLIP is shown in Figure \ref{Fig.data_generation}. It applies KOSMOS-2 and SAM to create diverse mask-caption pairs. First, the pipeline takes the input image and a text prompt, ``\textless grounding\textgreater Describe this image:", and fed to KOSMOS-2. The text prompt is given a \textless grounding\textgreater tag to guide KOSMOS-2 to locate image regions associated with texts in the output caption. This results in an image caption with box-caption pairs. In the second step, we convert the bounding boxes to binary masks by feeding the input image and each bounding box location to SAM. As a result, we are able to obtain diverse mask-caption pairs without any human intervention. \textbf{The final 1.3 million pairs of data would be released as the HumanCLIP dataset to facilitate future studies.} 

The pipeline shares a similar flow with PartSLIP++ to convert bounding boxes to binary masks with SAM. However, a significant advantage of our method is that \textbf{we do not assume text prompts for each mask are provided}. Instead, we adopt KOSMOS-2 to simultaneously generate box-caption pairs with a general text prompt. It ensures a more diverse and comprehensive coverage of contents within each image, enhancing the generalizability of our HumanCLIP model.


\begin{figure}[b]
    \centering
    \includegraphics[width=.95\hsize]{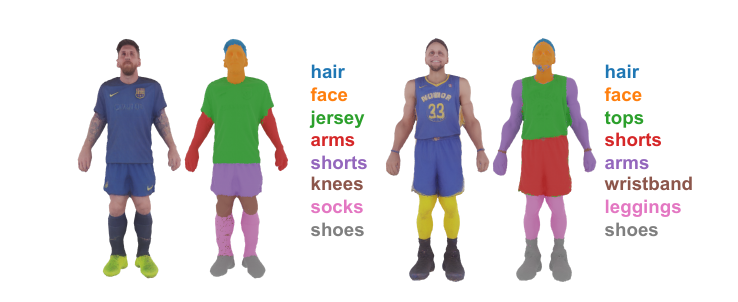}
    \caption{Segmentation of 3D humans generated from HumanNorm: an off-the-shelf text-to-3D model.} 
    \label{Fig.results_gen3d}
\end{figure}

\section{Few-shot Learning of PartSLIPs}
PartSLIP and PartSLIP++ are both claimed as low-shot methods, with their full potential unlocked through few-shot fine-tuning by category. We assess the performance of models that have undergone few-shot fine-tuning compared to the pre-trained checkpoints provided by the authors. We follow their low-shot setting and prepare point clouds to cover all of the part categories in the dataset with 8 samples as claimed in the paper. These are then rendered from 10 views to obtain images and ground truth bounding boxes. We keep the parameters of the GLIP model frozen and train only the learnable offset for each part. The few-shot results on the SIZER and CTD datasets are shown in Table \ref{tab:few_shot}. From the table, it can be observed that there is not a significant improvement in performance for both models on each dataset. Hence, we do not distinguish few-shot settings in our evaluations in the main paper.


\begin{table}
    \centering
    \caption{Few-shot learning efficacy of PartSLIP and PartSLIP++ on 3D human data.}
    \label{tab:few_shot}
    \resizebox{0.8\columnwidth}{!}{\begin{tabular}{c|cc|cc}
        \multirow{2}{*}{Model} & \multicolumn{2}{c|}{SIZER} & \multicolumn{2}{c}{CTD}\\ \cline{2-5}
        & Acc. & mIoU & Acc. & mIoU\\ \hline
         PartSLIP & 84.94 & 70.79 & 75.11 & 55.46 \\
         PartSLIP (few-shot) & 84.90 & 70.70 & 75.65 & 56.07 \\ \hline
         PartSLIP++& 86.63 & 72.93 & 80.73 & 62.36 \\ 
         PartSLIP++ (few-shot) & 86.67 & 72.99 & 80.69 & 62.49 
    \end{tabular}
    }
\end{table}

\section{More Qualitative Comparison}
\begin{figure*}[]
    \centering
    \includegraphics[width=\hsize]{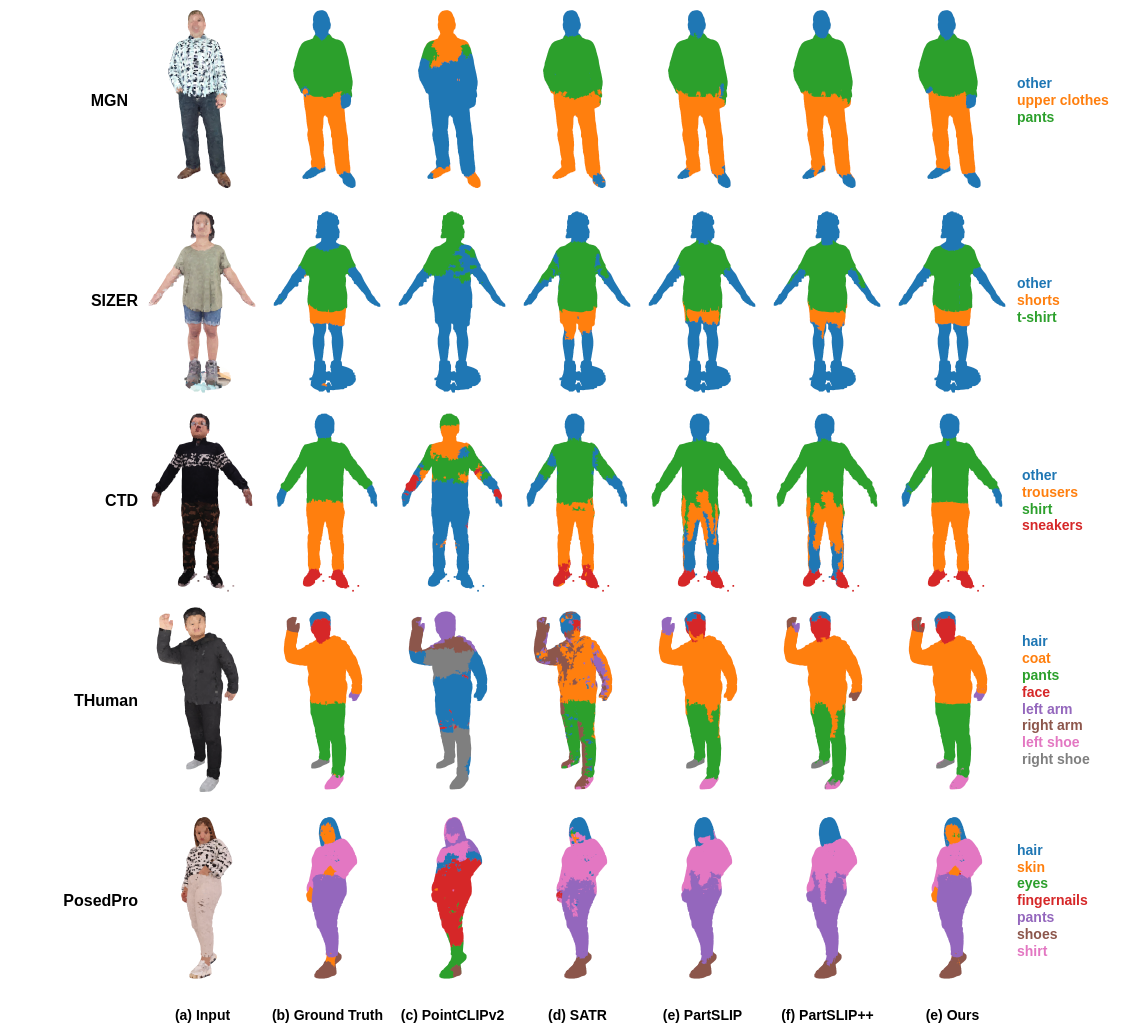}
    \caption{More Qualitative analysis of segmentation results with PointCLIPv2, SATR, PartSLIP, and PartSLIP++ on the 3D scans} 
    \label{Fig.visual_results_sup}
\end{figure*}

Additional visual comparisons for each dataset with other open-set 3D segmentation methods are shown in Figure \ref{Fig.visual_results_sup}.

\section{Segmentation of Generated 3D Human Models}
Due to the rapid development of 3D asset generation techniques, the models to be segmented may not originate from real-world scans. We demonstrate that our method effectively bridges the domain gap for less photorealistic generated data, significantly broadening its applicability across various content types. Figure \ref{Fig.results_gen3d} shows two examples of segmentation results on 3D humans generated from a text-to-3D human generation model. We use the output results from HumanNorm where the 3D humans were generated from the text descriptions: ``a DSLR photo of Messi" (left) and ``a DSLR photo of Stephen Curry" (right). In both cases, our framework can segment the generated models into distinct parts corresponding to the input prompts. It validates the robustness of our method to segment 3D humans at varying quality and content.

\section{Promptable Segmentation}
\begin{figure*}[]
    \centering
    \includegraphics[width=\hsize]{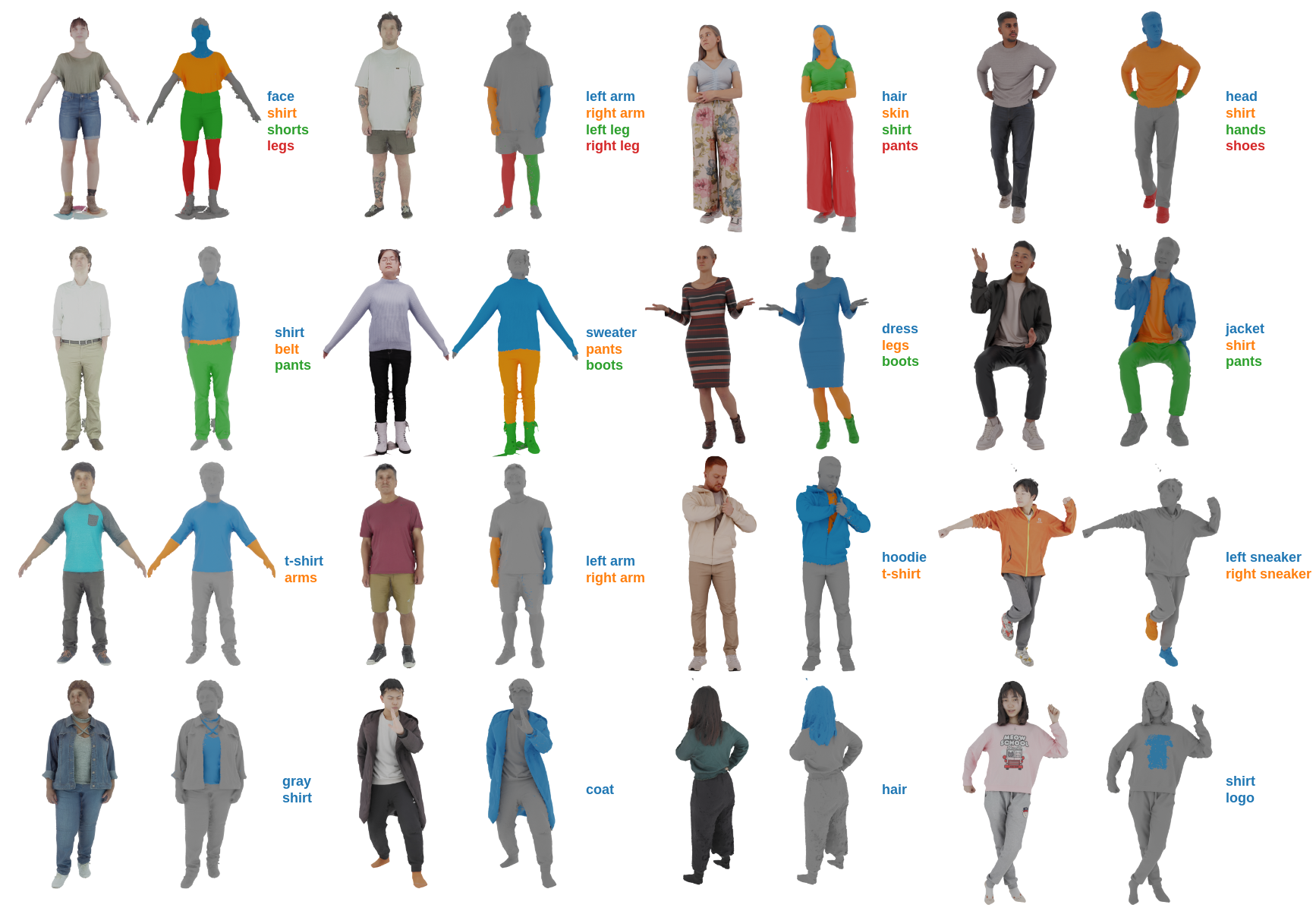}
    \caption{Additional examples of promptable segmentation. Each row shows a different type of combination of input prompts.} 
    \label{Fig.promptable_seg_sup}
\end{figure*}

Additional examples of our framework's promptable segmentation capability is visualized in Figure \ref{Fig.promptable_seg_sup}. In the figure, each row represents a different combination of input prompts to highlight our method's versatility in segmenting any category the user wants.

\begin{figure}[b]
    \centering
    \includegraphics[width=.95\hsize]{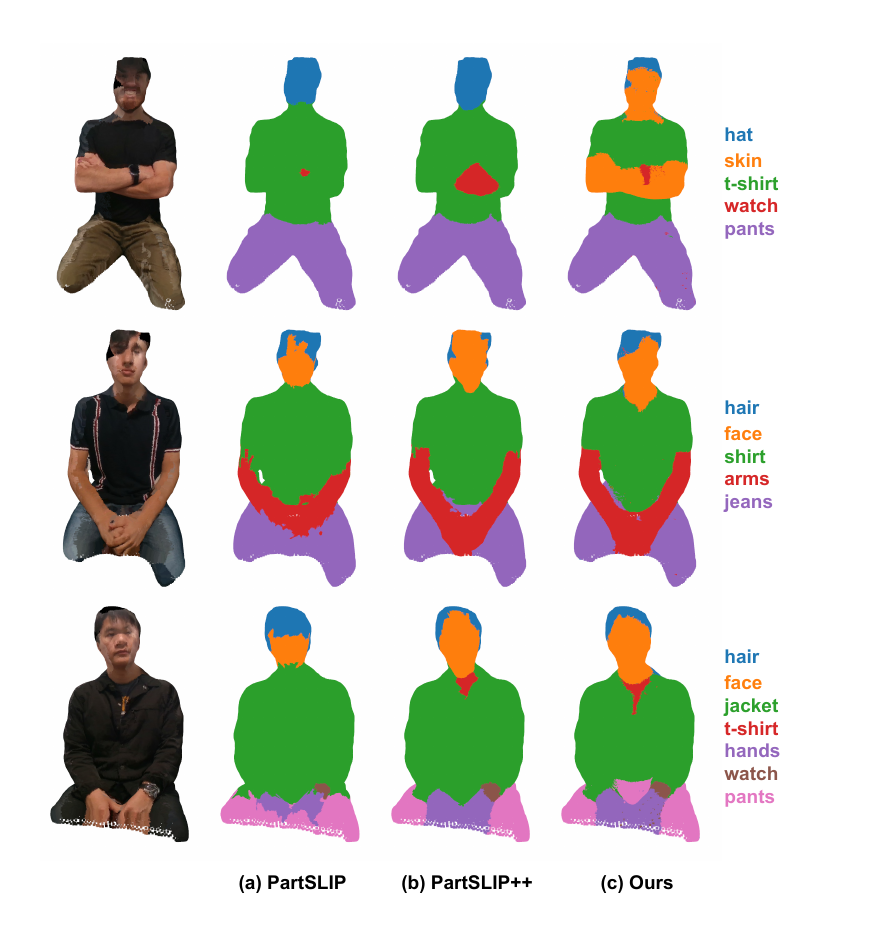}
    \caption{Additional visual comparisons of (a) PartSLIP, (b) PartSLIP++, and (c) Ours on our in-the-wild point cloud dataset.} 
    \label{Fig.results_wild_sup}
\end{figure}

\section{In-the-wild Segmentation}
Visual comparisons with PartSLIP and PartSLIP++ on our in-the-wild point dataset is shown in Figure \ref{Fig.results_wild_sup}.

\subsection{In-the-Wild Mesh Dataset Description}

We would release the 3D human mesh dataset we use under in-wild segmentation settings, consisting of 15 subjects in different poses to demonstrate different practical scenarios. We build a multi-view capturing system using consumer-level RGB-D cameras. To mimic the in-the-wild quality of 3D models, we utilize only four incomplete views and capture under the daylight environment, resulting in a relatively noisy and non-watertight surface. The 3D models are reconstructed through unprojecting and fusing. The meshes are created from fused point clouds through Poisson Surface Reconstruction. The generated meshes are then smoothed using a simple Laplacian filter to reduce high-frequency noise. \textbf{This dataset will be made open-source}, and contains the following poses and configurations:

\begin{itemize}
    \item \textbf{Sitting Pose:} Subjects were scanned in a natural sitting posture.
    \item \textbf{Arms Stretched Out:} Subjects were instructed to extend their arms fully, providing a clear view of the torso and limbs.
    \item \textbf{Human Object Interaction:} Subjects held various objects such as a bottle, book, or bag.
    \item \textbf{Loose Clothing:} Subjects wore loose clothing, such as hoodies.
    \item \textbf{Multi-Layer Clothing:} Subjects wore multiple visible layers of clothing.
\end{itemize}

\end{document}